\documentclass[11pt]{article}

\usepackage[final]{acl}

\usepackage{times}
\usepackage{latexsym}
\usepackage{makecell}
\usepackage{booktabs}   
\usepackage{multirow}
\usepackage{subcaption}
\usepackage{algorithm}
\usepackage{algpseudocode}
\usepackage{amsmath}
\usepackage[T1]{fontenc}

\usepackage[utf8]{inputenc}

\usepackage{microtype}

\usepackage{inconsolata}

\usepackage{graphicx}

\usepackage{amsmath}
\usepackage{comment}
\usepackage{dirtytalk}
\usepackage{multirow}
\usepackage{array}
\usepackage{amsfonts}
\usepackage{threeparttablex}

\title{Lightweight LLM Agent Memory with Small Language Models}

\author{
\textbf{Jiaquan Zhang\textsuperscript{1}},
\textbf{Chaoning Zhang\textsuperscript{1,*}},
\textbf{Shuxu Chen\textsuperscript{2}},
\textbf{Zhenzhen Huang\textsuperscript{1}},
\textbf{Pengcheng Zheng\textsuperscript{1}},\\
\textbf{Zhicheng Wang\textsuperscript{1}},
\textbf{Ping Guo\textsuperscript{3}},
\textbf{Fan Mo\textsuperscript{4}},
\textbf{Sung-Ho Bae\textsuperscript{2}},
\textbf{Jie Zou\textsuperscript{1}},
\textbf{Jiwei Wei\textsuperscript{1}},
\textbf{Yang Yang\textsuperscript{1}}\\
\textsuperscript{1}University of Electronic Science and Technology of China; \\
\textsuperscript{2}Kyung Hee University; 
\textsuperscript{3}City University of Hong Kong; 
\textsuperscript{4}University of Oxford\\
\small{\textbf{*Corresponding author:} \href{mailto:chaoningzhang1990@gmail.com}{chaoningzhang1990@gmail.com}}
}

\begin{document}
\maketitle
\begin{abstract}
Although LLM agents can leverage tools for complex tasks, they still need memory to maintain cross-turn consistency and accumulate reusable information in long-horizon interactions. However, retrieval-based external memory systems incur low online overhead but suffer from unstable accuracy due to limited query construction and candidate filtering. In contrast, many systems use repeated large-model calls for online memory operations, improving accuracy but accumulating latency over long interactions. We propose \textit{\textit{LightMem}}, a lightweight memory system for better agent memory driven by Small Language Models (SLMs). \textit{LightMem} modularizes memory retrieval, writing, and long-term consolidation, and separates online processing from offline consolidation to enable efficient memory invocation under bounded compute. We organize memory into short-term memory (STM) for immediate conversational context, mid-term memory (MTM) for reusable interaction summaries, and long-term memory (LTM) for consolidated knowledge, and uses user identifiers to support independent retrieval and incremental maintenance in multi-user settings. Online, \textit{LightMem} operates under a fixed retrieval budget and selects memories via a two-stage procedure: vector-based coarse retrieval followed by semantic consistency re-ranking. Offline, it abstracts reusable interaction evidence and incrementally integrates it into LTM. Experiments show consistent gains across model scales, with an average F1 improvement of about 2.5 over A-MEM on LoCoMo, while achieving higher efficiency and low median latency (83 ms for retrieval and 581 ms end-to-end).
\end{abstract}

\section{Introduction}
LLM-driven agents excel at long-term dialogue, multi-step reasoning, and task-oriented interaction \cite{zhang2026learning, huang2024understanding,zhu2025multiagentbench,zhang2026tdarctaskdrivenalignmentknowledgebased,zhang2026text,zhang2025spike}. To maintain cross-turn consistency beyond the context window, many systems augment agents with external memory \cite{DBLP:conf/icml/LeeCFCF24,xu2025mem,hu2025hiagent,wang2026efficient}. Long-term memory supports continual learning and planning.

\begin{figure}
    \centering
    \includegraphics[width=1\linewidth]{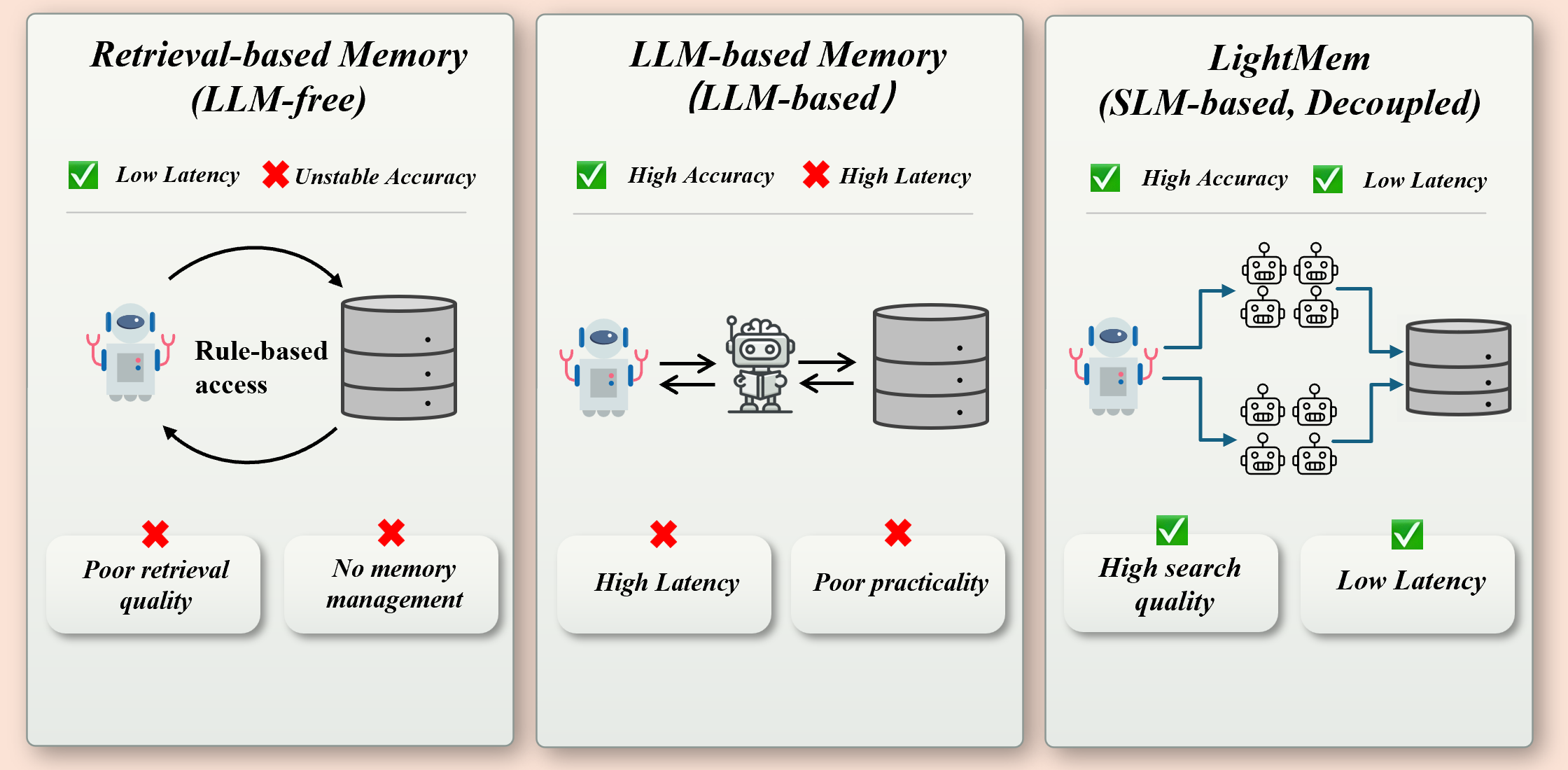}
    \caption{\textit{LightMem} combines enhanced retrieval with SLMs, achieving high retrieval accuracy while significantly reducing online latency compared to retrieval-based and LLM-based memory systems.}
    \label{fig:con}
\end{figure}
Existing memory systems can be broadly divided into two categories. One line is based on retrieval–driven external memory, where past interactions are compressed into retrievable entries \cite{asai2024self,DBLP:conf/acl/MaharanaLTBBF24} and the top-K memories are recalled via similarity search \cite{zhong2024memorybank}. These methods are efficient, but limited query construction and candidate filtering often introduce retrieval noise, resulting in unstable answer accuracy. Another line of methods \cite{DBLP:conf/acl/Tan000DD25,DBLP:journals/tois/ZhangDBMLCZDW25} introduces LLM-driven memory operations on top of external memory, repeatedly invoking large models for memory writing, retrieval and controlling. As these operations are typically implemented via repeated model invocations, they can introduce non-trivial runtime overhead over long interactions \cite{packer2023memgpt,park2023generative}. This contrast suggests a natural separation: keep high-frequency online memory decisions lightweight and controllable, while deferring heavy abstraction and consolidation to offline processing (see Figure \ref{fig:con}). Recent advances in SLMs \cite{magister2023teaching,han2024small} make this separation practical. SLMs can reliably handle high-frequency, structured decision tasks in online memory processing, enabling different memory stages to be assigned to models of appropriate scale \cite{liu2025smaller}. Online memory handling typically consists of structured subtasks (such as intent routing \cite{zhang2024mindmemory}, query construction \cite{hong2025enhancing} and semantic filtering \cite{hatalis2023memory}). These tasks place greater emphasis on predictable behavior and low overhead than on maximal generative capacity.
 In this setting, lightweight SLMs provide a suitable choice \cite{sinha2025small} for online control and filtering, while heavier abstraction and consolidation can be deferred to offline processing. Overall, existing memory systems face an efficiency–effectiveness trade-off. Moreover, SLMs are not a silver bullet due to limited capacity and representation. Bridging this gap requires strengthening the external memory pipeline, especially how memories are written and retrieved.








To address the above questions, we propose \textit{LightMem}, a lightweight memory system for LLM agents that models long-term memory as an incrementally evolvable process via specialized SLMs. \textit{LightMem} modularizes and decouples query parsing, memory retrieval, memory writing, and long-term consolidation, enabling independent optimization and elastic scaling of each component. This separation allows for lightweight online processing to be distinct from offline consolidation, achieving efficient memory use under tight compute budgets and supporting long-term evolution. Memories are organized into STM, MTM, and LTM stores based on temporal and access characteristics. User-identity metadata is embedded in each memory unit to enforce user-level logical isolation, balancing privacy, consistency, and scalability. We propose a modular online–offline memory pipeline in which distinct SLMs specialize in complementary memory operations. Online, LightMem uses three specialized small language model modules: a Controller (SLM-1) for intent and query planning, a Selector (SLM-2) for candidate verification and compression, and a Writer (SLM-3) for incremental memory writing. Specifically, the Controller rewrites the user input into intent-conditioned hypothetical queries (HQs) and allocates a fixed Top-$K$ budget across MTM and LTM. Given the retrieved candidates, the Selector performs metadata-constrained prefiltering followed by semantic-consistency based re-ranking to output the final Top-$K$ memories. After each turn, the Writer summarizes the interaction into compact MTM entries and maintains MTM incrementally. Long-term abstraction and consolidation are handled offline by a large-context model, keeping the online path lightweight. Our main contributions are summarized as: 
\begin{itemize}
    \item We propose \textit{LightMem}, a lightweight memory system collaboratively driven by SLMs. Different SLMs handle lightweight, high-frequency online memory operations (query construction, retrieval, and writing), while heavier abstraction and consolidation are deferred to offline processing.
    \item We propose a two-stage memory querying design. Given a user query, \textit{LightMem} first narrows down the candidate set with fast retrieval, and then applies semantic-level verification to select the truly relevant memories.
    \item We evaluate \textit{LightMem} on LoCoMo and DialSim, demonstrating gains across model scales, including an average F1 improvement of about 2.5 on LoCoMo, improved semantic consistency on DialSim and low median latency (83 ms retrieval; 581 ms end-to-end).
\end{itemize}

\section{Related Work}
\subsection{LLM Memory Systems}
Prior work can be broadly categorized into retrieval-based memory and LLM-driven memory operations. \textbf{Retrieval-based Memory.} MemoryBank \cite{zhong2024memorybank} supports personalized long-term dialogue by storing summarized user events in an external memory and retrieving relevant entries for each query, with forgetting to control growth. MemGPT \cite{packer2023memgpt} treats the context window as virtual memory and performs paging between the prompt and an external store, with runtime eviction and on-demand retrieval. ReadAgent \cite{DBLP:conf/icml/LeeCFCF24} similarly relies on retrieval over an external cache: it indexes compressed gists and performs on-demand lookup to fetch supporting evidence when needed. \textbf{LLM-driven Memory.}
HiAgent \cite{hu2025hiagent} manages in-trial working memory by chunking trajectories into subgoals and summarizing past steps with LLMs. A-MEM \cite{xu2025mem} further builds self-organizing memory networks through LLM-driven note-taking and automatic linking, but typically does not emphasize strict online/offline decoupling under resource constraints. 
These lines expose a recurring trade-off between lightweight but noisy retrieval-based memory and more effective yet costly LLM-driven online memory operations. \textit{LightMem} addresses it with SLM-based lightweight online control and filtering, coupled with improved external memory writing and retrieval under a fixed budget.

\section{Method}
\label{sec:method}

\begin{figure*}[h]
    \centering
    \includegraphics[width=0.9\linewidth]{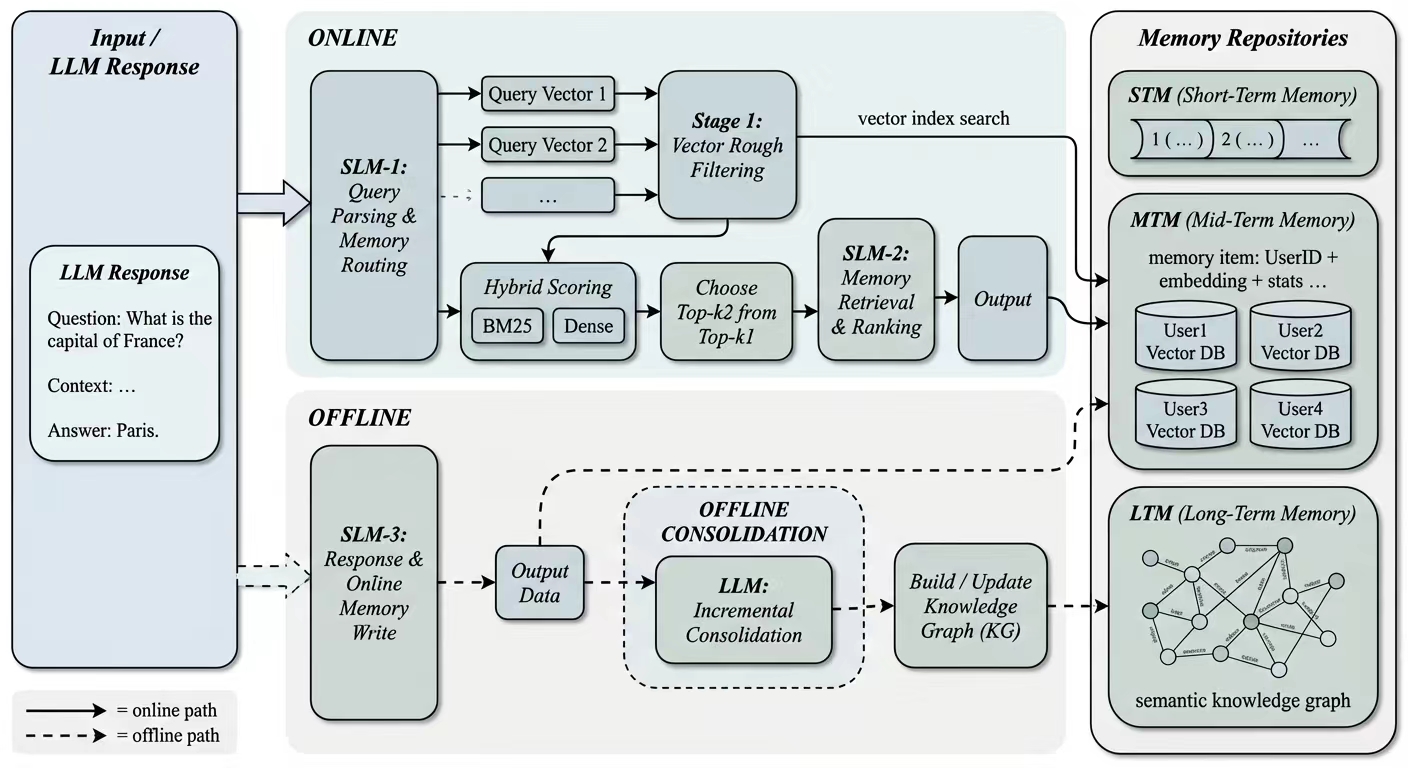}
    \caption{Multiple SLMs coordinate an online pathway for query-time routing and retrieval over STM/MTM, and an offline pathway that incrementally consolidates MTM into a graph-structured LTM.}
    \label{fig:1}
\end{figure*}

As shown in Figure~\ref{fig:1}, \textit{LightMem} uses specialized SLMs to modularize memory operations, separating lightweight online querying from offline long-term consolidation.


\subsection{Problem Setup and Preliminaries}
\label{se1}

We consider a multi-turn dialogue setting with multiple users. At turn $t$, a user provides an input $x_t$, and the model generates a response $y_t$. Due to the limited context capacity of SLMs, the accessible dialogue context at turn $t$ is denoted by $C_t$, typically a truncated window of recent turns. \textit{LightMem} maintains a user-scoped memory store $M_u$ with user-level logical isolation via user identifiers (see \S\ref{sec:memory_stores} for the definition of STM/MTM/LTM). Given $(x_t, C_t)$, the system retrieves a memory set $R_t \subseteq M_u$ with $|R_t|\le K$ and uses $R_t$ to assist in generating $y_t$. The specific symbols are provided in the Appendix \ref{ap1}.

\subsection{Memory Stores}
\label{sec:memory_stores}

\paragraph{STM.}
STM is implemented as the SLM context window that buffers the most recent interaction sequence in the current session. It serves only as working memory and is updated turn by turn within the prompt; STM is neither persisted nor retrieved.

\paragraph{MTM.}
MTM is the sole carrier of personalized episodic memory. It stores a set of memory items, each consisting of (i) a concise semantic summary, (ii) temporal information and access statistics, (iii) an embedding for similarity-based retrieval, and (iv) a user identifier for strict per-user isolation.

\paragraph{LTM.}
LTM stores de-identified, user-agnostic semantic knowledge distilled offline from high-value MTM episodes. It contains no raw personal episodes or user identifiers; instead, it represents stable facts, domain regularities, and cross-user trends. LTM is organized as a lightweight graph-structured knowledge base to support multi-hop reasoning and knowledge sharing across tasks.

\subsection{Intent Modeling and Retrieval Control}
\label{se2}

Given $(x_t, C_t)$, SLM-1 serves as a lightweight retrieval controller rather than a retriever. It converts the raw input into a structured retrieval plan that specifies: (i) what to search (queries), (ii) how to constrain the search (metadata filters), and (iii) how many items to return under a fixed Top-$K$ budget. Concretely, SLM-1 first infers coarse intent attributes, e.g., whether the query depends more on recent episodic details or long-term stable knowledge, and whether strong personalization is needed. These attributes are used only for retrieval planning (query rewriting, routing, and budgeting), not for answer generation. SLM-1 then rewrites the input into a set of HQs $\{q_t^{(i)}\}$ and outputs metadata constraints $\phi_t$ (user identifier, optional time window, and type tags) to reduce noise and enforce user-level isolation. Finally, it issues the retrieval request:
\begin{equation}
\mathcal{Q}_t = \left\langle \{q_t^{(i)}\},\ \phi_t,\ K \right\rangle.
\end{equation}
This request is consumed by \S\ref{se3}.




\subsection{Two-Stage Retrieval}
\label{se3}

Given $\mathcal{Q}_t$, SLM-2 executes a two-stage retrieval procedure and returns a memory set $R_t$ that satisfies the Top-$K$ constraint:
\begin{equation}
R_t = \textsc{Retrieve}(\mathcal{Q}_t), \qquad |R_t| \le K.
\end{equation}
The retrieved memories $R_t$ are appended to the context to condition response generation for $y_t$.

For a single user query, SLM-1 may generate $n$ HQs, denoted by $\{q_t^{(i)}\}_{i=1}^{n}$, to improve recall coverage. We impose a single final return budget $K$ and adopt a double-budget rule in the coarse stage: the total number of candidates returned by Stage~1 is fixed to $2K$.

\paragraph{Stage 1: metadata-constrained coarse retrieval.}
Under metadata constraints $\phi_t$ (including the user identifier, an optional time window, and type/source tags), the system performs vector-based coarse retrieval for each HQ. The Stage~1 candidate budget is evenly split across HQs:
\begin{equation}
\sum_{i=1}^{n} K_1^{(i)} = 2K, 
\qquad
K_1^{(i)} = \frac{2K}{n}.
\end{equation}
Let $C^{(i)}$ denote the candidate list returned for the $i$-th HQ. The aggregated candidate set is
\begin{equation}
C = \bigcup_{i=1}^{n} C^{(i)}, \qquad |C| = 2K.
\end{equation}
Stage~1 is designed for coverage, acting as an efficient filter that compresses the large search space into a fixed-size candidate set.

\paragraph{Stage 2: semantic filtering and compression.}
In Stage~2, SLM-2 takes the HQ set $\{q_t^{(i)}\}_{i=1}^{n}$ and the Stage~1 candidates $C$ (memory summaries with necessary structured metadata) and performs semantic consistency checking and relevance judgment. Instead of generating answers, SLM-2 conducts controlled selection on a fixed-size candidate pool: it keeps the most relevant half of $|C|=2K$ candidates and outputs the final retrieval results
\begin{equation}
R_t \subseteq C, \qquad |R_t| \le K
\end{equation}
This two-to-one compression yields (i) stable computation with a fixed candidate size, (ii) semantic refinement beyond vector similarity by leveraging intent and metadata, and (iii) noise suppression by explicitly discarding roughly half of the candidates. The resulting Top-$K$ memories $R_t$ are provided to the downstream agent or response generator.

\subsection{Memory Writing and Update}
\label{se4}
After generating $y_t$, SLM-3 extracts reusable user-relevant information from the current interaction, compresses it into a concise memory item, and appends it to MTM. To prevent redundancy and noise accumulation, highly repetitive items are merged or rewritten, while conflicting information is handled using temporal cues and evidence strength. To ensure consistently low-latency personalized retrieval, we enforce a capacity bound on MTM:
\begin{equation}
|M_u^{\textsc{MTM}}| \le B
\end{equation}
When the bound is reached, MTM is maintained by evicting stale, low-utility items and further compressing redundant content.

\subsection{Offline Consolidation}
\label{se5}

To avoid increasing online retrieval and writing latency, \textit{LightMem} performs offline consolidation to distill high-value episodic evidence in MTM into de-identified, long-term semantic knowledge, enabling sustained evolution of LTM. An LLM handles this offline path with a large context window and is strictly decoupled from online operation. In each cycle, the LLM processes only an incremental batch (newly written or retrieval-reactivated MTM items and low-utility candidates flagged under MTM capacity pressure), rather than rebuilding MTM/LTM from scratch. It abstracts episodes into privacy-preserving knowledge candidates, performs similarity search over LTM to locate semantically nearest anchors, and incrementally inserts and links candidates within the local neighborhood, thereby maintaining LTM as lightweight graph-structured knowledge. Evidence accumulated over time further drives merge/update/drop decisions, while confidence decay is applied to weakly supported candidates to enable natural forgetting and to limit the impact of stale or incidental information on subsequent retrieval and reasoning. We provide each component with algorithmic details deferred to Appendix~\ref{ap1}.

\section{Experiment}
\subsection{Experimental Setup} 
\subsubsection{Datasets}

To evaluate the effectiveness of \textit{\textit{LightMem}}, we conduct experiments on two datasets. LoCoMo \cite{DBLP:conf/acl/MaharanaLTBBF24} is a benchmark designed to evaluate logical reasoning over extended conversational contexts. It contains long dialogues (on average 9K tokens) and covers five core task categories: Single-hop, Multi-hop, Temporal, Open-domain, and Adversarial. 
DialSim \cite{kim2024dialsim} is a dialogue simulation dataset derived from TV shows, containing 1,300 multi-party sessions spanning years.

\subsubsection{Baselines}
\textbf{Baselines} We select baselines that are representative in prior work on long-term memory, covering common designs for storing, retrieving, and updating dialogue histories and that can be evaluated under the same backbone and retrieval budget for a fair comparison: LoCoMo \cite{DBLP:conf/acl/MaharanaLTBBF24}, ReadAgent \cite{DBLP:conf/icml/LeeCFCF24}, MemoryBank \cite{zhong2024memorybank}, MemGPT \cite{packer2023memgpt}, and A-MEM \cite{xu2025mem}. When selecting backbone models, a diverse array of LLMs is used to assess their generalization capabilities. These models encompass GPT-4o, GPT-4o-mini \cite{hurst2024gpt}, Qwen2.5 (available in 1.5B and 3B) \cite{bai2025qwen2}, and Llama 3.2 (offered in 1B and 3B) \cite{dubey2024llama}.

\subsubsection{Metric}
We evaluate response quality using both lexical-overlap and semantic metrics. Specifically, we report F1 and BLEU-1 to measure answer correctness and token-level overlap with reference responses, which are commonly used for short factual and span-like outputs. For DialSim \cite{kim2024dialsim}, where multiple surface forms can express the same meaning, we additionally report ROUGE-L and METEOR to capture sequence-level overlap and soft token matching, and use SBERT similarity to quantify semantic consistency between generated and reference responses beyond n-gram matches. These metrics provide a reliable assessment of both precision and meaning retention in long-horizon dialogue scenarios.

\subsubsection{Implement Details}
To isolate the effect of memory, we use GPT-4o-mini as the fixed response generator for \textit{\textit{LightMem}} and all baselines. \textit{\textit{LightMem}}'s control plane uses locally deployed, quantized SLMs: SLM-1 for HQ generation and retrieval control, SLM-2 for semantic re-ranking and compression, and SLM-3 for online writing and MTM maintenance. For HQ generation, SLM-1 follows a structured prompt for query decomposition and routing; Table~\ref{tab:hq_prompt} summarizes the prompt logic and provides an example.
We deploy quantized Llama-3.2-1B-Instruct via Ollama by default and additionally evaluate Qwen2.5-1.5B-Instruct for robustness. Offline consolidation is handled by a large-context LLM and is decoupled from the online path.
We encode all queries and memory entries using all-MiniLM-L6-v2 (384 dimensions) and retrieve the top-10 nearest neighbors in the coarse vector retrieval stage. SLM-2 is fine-tuned with LoRA on 2{,}000 constructed (Query, Subgraph, Path) samples. We cap MTM capacity at $B=10^4$ and apply dynamic pruning beyond this limit by evicting stale, low-utility items and merging/compressing redundant content. \textit{\textit{LightMem}} inference and latency measurements are conducted on a single NVIDIA RTX 4090 (24GB) GPU. Other unspecified settings follow A-MEM \cite{xu2025mem}.

\begin{table}[h]
\centering
\caption{Structured HQ prompt used by SLM-1 for query decomposition and routing.}
\small
\setlength{\tabcolsep}{6pt}
\begin{tabular}{p{0.20\linewidth} p{0.72\linewidth}}
\toprule
\textbf{HQ prompt step} & \textbf{ Example} \\
\midrule
Detect missing info &
Identify underspecified references (e.g., pronouns such as ``it/that/he/the project''), implicit context dependencies, and vague time cues (e.g., ``recently/last time/before''). \\
\addlinespace
Generate HQs &
Rewrite the user request into one or more standalone HQs, optionally splitting intent into user-specific vs.\ factual queries, and making implied context (time/location) explicit. \\
\addlinespace
Route and budget &
Assign each HQ to MTM (user-specific) or LTM (general/public knowledge) and allocate a retrieval budget under a fixed top-$K$ constraint. \\
\midrule
Example &
Input: ``Recommend a dinner spot.'' The prompt yields two HQs: one asks about the user's preferred cuisines or dietary constraints (routed to MTM), and the other asks about highly rated nearby restaurants (routed to LTM). \\
\bottomrule
\end{tabular}

\label{tab:hq_prompt}
\end{table}

\begin{table*}[h]
\centering
\caption{Main results on LoCoMo across question categories. We report F1 and BLEU-1 for each category, along with the token length of the effective context. Best results within each model block are in bold.}
\label{tab:main_locomo_results}
\setlength{\tabcolsep}{3.2pt}
\renewcommand{\arraystretch}{1.05}
\scriptsize
\resizebox{\linewidth}{!}{%
\begin{tabular}{llccccccccccc}
\toprule
Model & Method
& \multicolumn{2}{c}{Single-hop}
& \multicolumn{2}{c}{Multi-hop}
& \multicolumn{2}{c}{Temporal}
& \multicolumn{2}{c}{Open-domain}
& \multicolumn{2}{c}{Adversarial}
& Token Length \\
\cmidrule(lr){3-4}\cmidrule(lr){5-6}\cmidrule(lr){7-8}\cmidrule(lr){9-10}\cmidrule(lr){11-12}
& & F1 & BLEU & F1 & BLEU & F1 & BLEU & F1 & BLEU & F1 & BLEU & \\
\midrule

\multirow{6}{*}{GPT-4o-mini}
& LoCoMo     & 40.36 & 29.05 & 25.02 & 19.75 & 18.41 & 14.77 & 12.04 & 11.16 & \textbf{69.23} & \textbf{68.75} & 16{,}910 \\
& ReadAgent  &  9.67 &  7.66 &  9.15 &  6.48 & 12.60 &  8.87 &  5.31 &  5.12 &  9.81 &  9.02 &    643 \\
& MemoryBank &  6.61 &  5.16 &  5.00 &  4.77 &  9.68 &  6.99 &  5.56 &  5.94 &  7.36 &  6.48 &    432 \\
& MemGPT     & 41.04 & 34.34 & 26.65 & 17.72 & 25.52 & 19.44 &  9.15 &  7.44 & 43.29 & 42.73 & 16{,}977 \\
& A-MEM      & 44.65 & 37.06 & 27.02 & 20.09 & 45.85 & 36.67 & 12.14 & 12.00 & 50.03 & 49.47 &  2{,}520 \\
& LightMem   & \textbf{45.81} & \textbf{38.24} & \textbf{28.85} & \textbf{21.43} & \textbf{46.28} & \textbf{37.15} & \textbf{13.52} & \textbf{12.96} & 54.57 & 52.19 &  1{,}150 \\
\midrule

\multirow{6}{*}{GPT-4o}
& LoCoMo     & \textbf{61.56} & \textbf{54.19} & 28.00 & 18.47 &  9.09 &  5.78 & 16.47 & 14.80 & \textbf{52.61} & \textbf{51.13} & 16{,}910 \\
& ReadAgent  & 12.46 & 10.29 & 14.61 &  9.95 &  4.16 &  3.19 &  8.84 &  8.37 &  6.81 &  6.13 &    805 \\
& MemoryBank &  8.28 &  7.10 &  6.49 &  4.69 &  2.47 &  2.43 &  6.43 &  5.30 &  4.42 &  3.67 &    569 \\
& MemGPT     & 60.16 & 53.35 & 30.36 & 22.83 & 17.29 & 13.18 & 12.24 & 11.87 & 34.96 & 34.25 & 16{,}987 \\
& A-MEM      & 48.43 & 42.97 & 32.86 & 23.76 & 39.41 & 31.23 & 17.10 & 15.84 & 36.35 & 35.53 &  1{,}216 \\
& LightMem   & 49.87 & 44.25 & \textbf{34.52} & \textbf{25.18} & \textbf{40.16} & \textbf{32.04} & \textbf{18.43} & \textbf{16.91} & 41.59 & 39.82 &    685 \\
\midrule

\multirow{6}{*}{Qwen2.5-1.5B}
& LoCoMo     & 11.15 &  8.67 &  9.05 &  6.55 &  4.25 &  4.04 &  9.91 &  8.50 & 40.38 & 40.23 & 16{,}910 \\
& ReadAgent  & 10.13 &  7.54 &  6.61 &  4.93 &  2.55 &  2.51 &  5.31 & 12.24 &  5.42 & 27.32 &    752 \\
& MemoryBank & 13.42 & 11.01 & 11.14 &  8.25 &  4.46 &  2.87 &  8.05 &  6.21 & 36.76 & 34.00 &    284 \\
& MemGPT     &  9.56 &  7.34 & 10.44 &  7.61 &  4.21 &  3.89 & 13.42 & 11.64 & 31.51 & 28.90 & 16{,}953 \\
& A-MEM      & 23.63 & 19.23 & 18.23 & 11.94 & 24.32 & 19.74 & 16.48 & 14.31 & 46.00 & 43.26 &  1{,}300 \\
& LightMem   & \textbf{25.13} & \textbf{20.56} & \textbf{21.57} & \textbf{13.82} & \textbf{25.64} & \textbf{20.89} & \textbf{17.21} & \textbf{15.18} & \textbf{49.82} & \textbf{46.54} &    728 \\
\midrule

\multirow{6}{*}{Qwen2.5-3B}
& LoCoMo     &  7.03 &  5.69 &  4.61 &  4.29 &  3.11 &  2.71 &  4.55 &  5.97 & 16.95 & 14.81 & 16{,}910 \\
& ReadAgent  &  3.25 &  2.51 &  2.47 &  1.78 &  3.01 &  3.01 &  5.57 &  5.22 & 15.78 & 14.01 &    776 \\
& MemoryBank &  4.11 &  3.32 &  3.60 &  3.39 &  1.72 &  1.97 &  6.63 &  6.58 & 13.07 & 10.30 &    298 \\
& MemGPT     &  7.26 &  5.52 &  5.07 &  4.31 &  2.94 &  2.95 &  7.04 &  7.10 & 14.47 & 12.39 & 16{,}961 \\
& A-MEM      & 17.23 & 13.12 & 12.57 &  9.01 & 27.59 & 25.07 &  7.12 &  7.28 & 27.91 & 25.15 &  1{,}137 \\
& LightMem   & \textbf{19.85} & \textbf{15.21} & \textbf{15.86} & \textbf{11.23} & \textbf{29.17} & \textbf{26.52} & \textbf{9.58} & \textbf{9.14} & \textbf{33.54} & \textbf{29.87} &    611 \\
\midrule

\multirow{6}{*}{Llama-3.2-1B}
& LoCoMo     & 12.86 & 10.50 & 11.25 &  9.18 &  7.38 &  6.82 & 11.90 & 10.38 & 51.89 & 48.27 & 16{,}910 \\
& ReadAgent  &  7.75 &  6.03 &  5.96 &  5.12 &  1.93 &  2.30 & 12.46 & 11.17 & 44.64 & 40.15 &    665 \\
& MemoryBank & 17.30 & 14.03 & 13.18 & 10.03 &  7.61 &  6.27 & 15.78 & 12.94 & 52.61 & 47.53 &    274 \\
& MemGPT     & 10.16 &  7.68 &  9.19 &  6.96 &  4.02 &  4.79 & 11.14 &  8.24 & 49.75 & 45.11 & 16{,}950 \\
& A-MEM      & 28.51 & 24.13 & 19.06 & 11.71 & 17.80 & 10.28 & 17.55 & 14.67 & 58.81 & 54.28 &  1{,}376 \\
& LightMem   & \textbf{30.88} & \textbf{26.24} & \textbf{22.41} & \textbf{14.27} & \textbf{19.93} & \textbf{12.16} & \textbf{18.95} & \textbf{16.12} & \textbf{63.52} & \textbf{58.49} &    754 \\
\midrule

\multirow{6}{*}{Llama-3.2-3B}
& LoCoMo     &  8.37 &  6.93 &  6.88 &  5.77 &  4.37 &  4.40 & 10.65 &  9.29 & 30.25 & 28.46 & 16{,}910 \\
& ReadAgent  &  3.25 &  2.51 &  2.47 &  1.78 &  3.01 &  3.01 &  5.57 &  5.22 & 15.78 & 14.01 &    461 \\
& MemoryBank &  7.61 &  6.03 &  6.19 &  4.47 &  3.49 &  3.13 &  4.07 &  4.57 & 18.65 & 17.05 &    263 \\
& MemGPT     &  4.32 &  3.51 &  5.32 &  3.99 &  2.68 &  2.72 &  5.64 &  5.54 & 21.45 & 19.37 & 16{,}956 \\
& A-MEM      & 28.14 & 23.87 & 17.44 & 11.74 & 26.38 & 19.50 & 12.53 & 11.83 & 42.04 & 40.60 &  1{,}126 \\
& LightMem   & \textbf{30.26} & \textbf{25.91} & \textbf{20.15} & \textbf{13.54} & \textbf{28.29} & \textbf{21.13} & \textbf{14.28} & \textbf{13.57} & \textbf{47.16} & \textbf{44.83} &    642 \\
\bottomrule
\end{tabular}%
}
\end{table*}

\subsection{Performance}
\textbf{Performance.} As shown in Table~\ref{tab:main_locomo_results}, \textit{\textit{LightMem}} achieves the best overall performance on the LoCoMo benchmark. In comparison to baselines that directly rely on long-context replay (such as LoCoMo and MemGPT), \textit{\textit{LightMem}} demonstrates more stable performance across the majority of question categories. It has particularly distinct advantages for multi-hop and temporal questions, all while not relying on a single long-context input. In comparison with summarization- or compression-based methods (MemoryBank and ReadAgent), \textit{\textit{LightMem}} consistently outperforms them across all categories, with larger margins on multi-hop and temporal reasoning. When compared to the strongest memory-centric baseline, A-MEM, \textit{\textit{LightMem}} further improves overall performance in most model settings. On GPT-4o, \textit{\textit{LightMem}} attains an F1 score of 34.52 for multi-hop questions, surpassing A-MEM's score of 32.86. Overall, \textit{\textit{LightMem}} demonstrates more robust long-term memory across different model scales and question types. In addition, it maintains significant advantages in cross-session reasoning tasks.

Table~\ref{tab:dialsim_comparison} reports the comparison on DialSim with GPT-4o-mini. We incorporate METEOR and SBERT similarity. This is driven by the characteristics of long-term dialogue generation. In long-term dialogue generation, semantically accurate answers can be presented in diverse surface forms. Consequently, metrics based solely on n-grams may undervalue advancements. METEOR partially accounts for morphological and synonym-level matching, while SBERT directly measures semantic alignment between the generated response and the reference, providing a complementary view of memory effectiveness. Across all metrics, \textit{\textit{LightMem}} achieves the best results. Importantly, \textit{\textit{LightMem}} improves not only lexical overlap (e.g., ROUGE-L/ROUGE-2 and METEOR) but also semantic similarity, with SBERT increasing from 19.51 (A-MEM) to 23.40. This indicates that the gains are not limited to reproducing similar wording, but also reflect stronger semantic consistency enabled by more effective use of long-term conversational history.

\begin{table}[h]
\centering
\caption{Comparison of different memory mechanisms on DialSim with GPT-4o-mini.}
\label{tab:dialsim_comparison}

\resizebox{\linewidth}{!}{%
\begin{tabular}{lcccccc}
\toprule
Method & F1 & BLEU-1 & ROUGE-L & ROUGE-2 & METEOR & SBERT  \\
\midrule
LoCoMo        & 2.55 & 3.13 & 2.75 & 0.90 & 1.64 & 15.76 \\
MemGPT        & 1.18 & 1.07 & 0.96 & 0.42 & 0.95 & 8.54 \\
A-MEM         & 3.45 & 3.37 & 3.54 & 3.60 & 2.05 & 19.51 \\
LightMem & \textbf{4.12} & \textbf{3.95} & \textbf{4.20} & \textbf{4.15} & \textbf{2.48} & \textbf{23.40} \\
\bottomrule
\end{tabular}}
\end{table}

\noindent\textbf{Generalization.} We evaluate all methods across diverse backbones, ranging from large models (GPT-4o / GPT-4o-mini) to small open-source models (Qwen2.5 and Llama-3.2). As shown in Table~\ref{tab:main_locomo_results}, \textit{\textit{LightMem}} remains the best or near-best method across these settings, indicating that its gains are robust and not tied to a specific backbone.

\subsection{Ablation Study}

To quantify the contribution of each component in \textit{\textit{LightMem}}, we conduct ablation studies by removing or simplifying one key module at a time while keeping the rest of the system unchanged. We consider the following variants:

\begin{itemize}
    \item \textbf{w/o semantic reranking} We remove the LM-based semantic filtering stage and keep only embedding-based Top-$K$ retrieval.
    \item \textbf{w/o HQ and retrieval routing} We disable HQs as well as the controller-based routing/budget allocation, and retrieve using the original query directly.
    \item \textbf{w/o MTM} We remove the MTM layer, relying only on the current context (STM) and LTM for retrieval and generation.
    \item \textbf{w/o offline consolidation} We disable the offline consolidation pipeline so that LTM is not periodically updated from MTM.
    \item \textbf{w/o graph structure} We replace the graph-structured LTM with a flat vector store.
\end{itemize}


\begin{figure*}[t]
    \centering
    \begin{minipage}{0.49\linewidth}
        \centering
        
        \includegraphics[width=\linewidth]{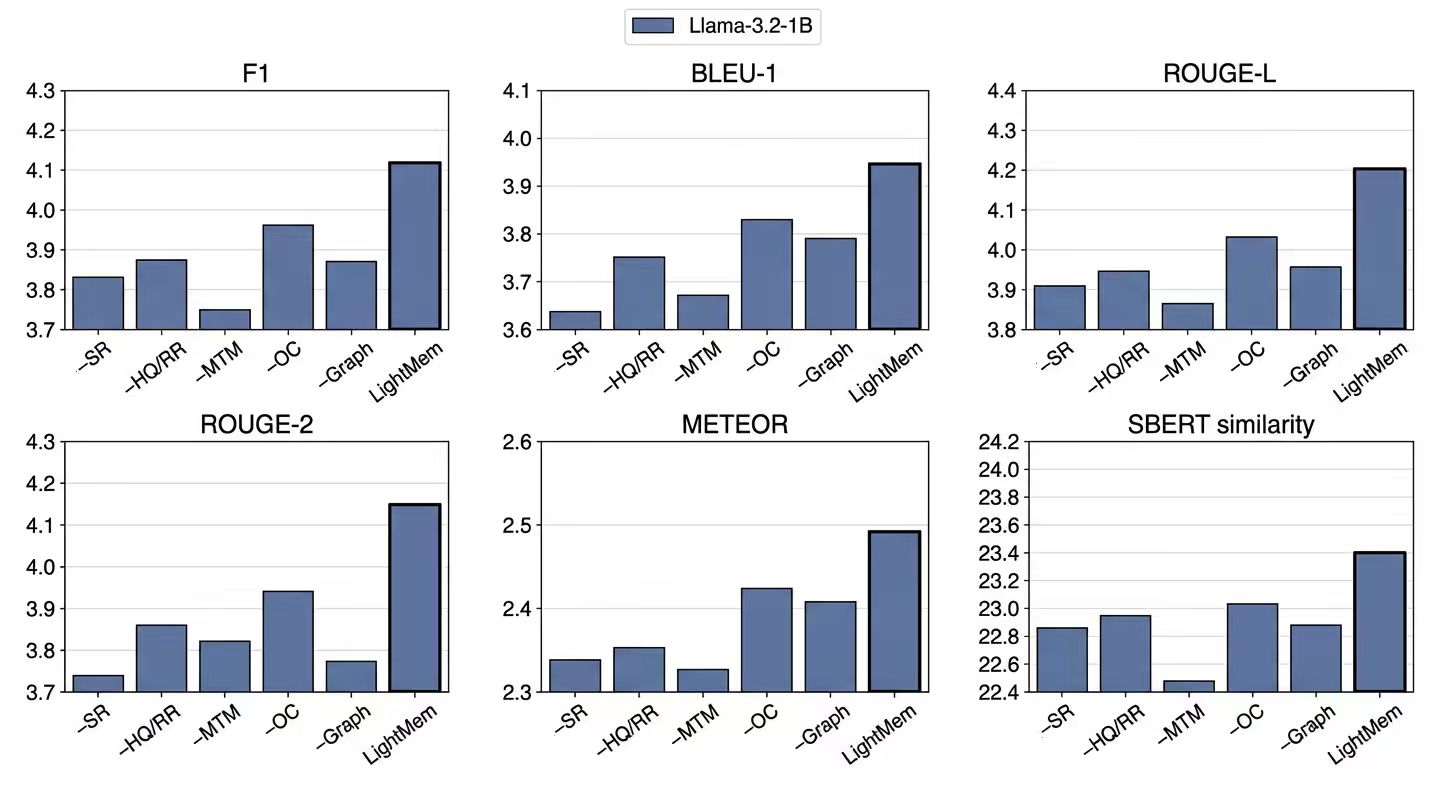}
        \subcaption{Ablation study on Dialsim with Llama-3.2-1B}
    \end{minipage}
    \hfill
    \begin{minipage}{0.49\linewidth}
        \centering
        \includegraphics[width=\linewidth]{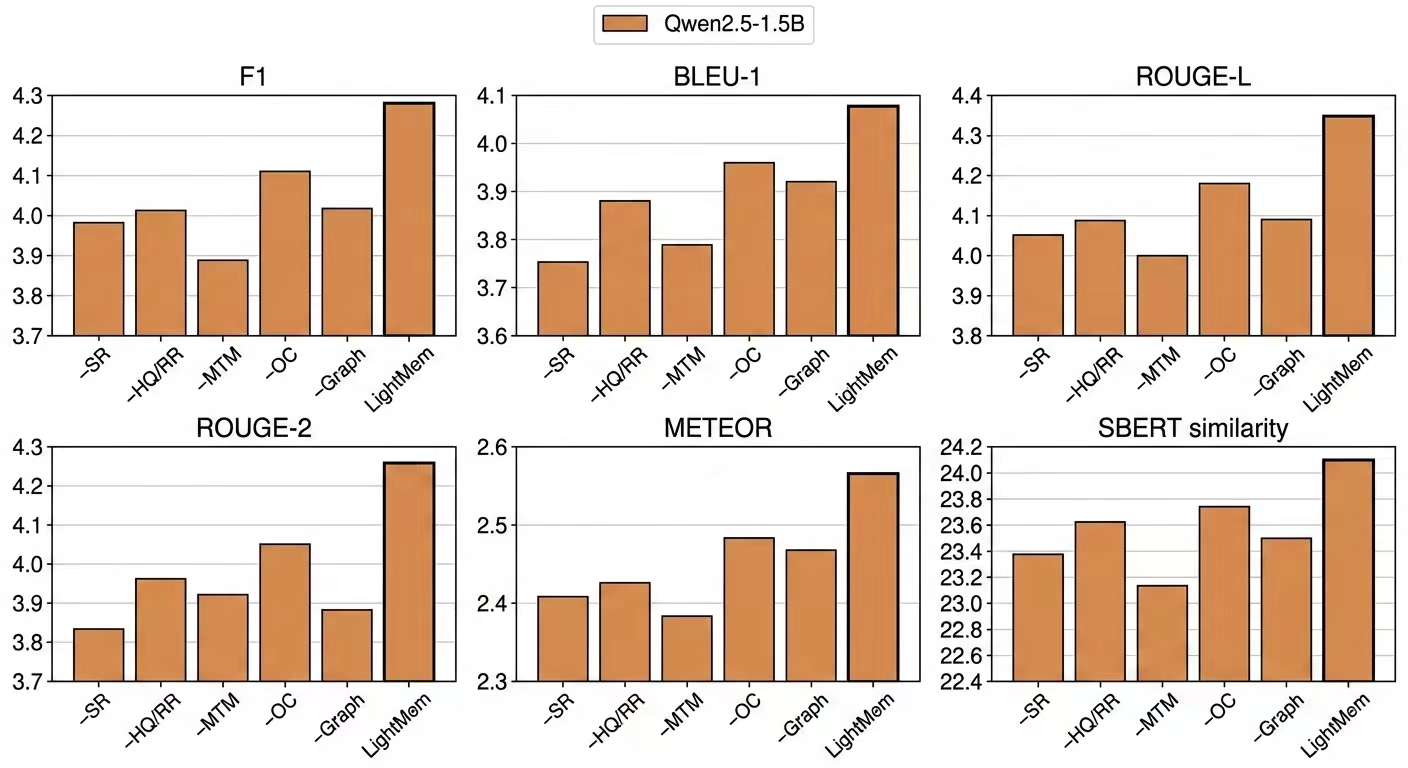}
        \subcaption{Ablation study on Dialsim with Qwen2.5-1.5B}
    \end{minipage}
    \caption{Ablation study on DialSim. We report F1, BLEU-1, ROUGE-L, ROUGE-2, METEOR, and SBERT similarity.}
    \label{fig:combined_bars}
\end{figure*}

Figure~\ref{fig:combined_bars} shows that removing any single component consistently degrades performance across metrics. Disabling semantic reranking or HQ-based retrieval routing leads to clear drops in F1; on Llama-3.2-1B, \textit{\textit{LightMem}} achieves an F1 of 4.12, while the variant without semantic reranking scores 3.83 and the variant without HQ and retrieval routing scores 3.87. Removing MTM further lowers performance, with an F1 of 3.75, indicating the benefit of retaining mid-term episodic memory. 
In comparison, disabling offline consolidation results in a smaller but still consistent decline, with an F1 of 3.96 on Llama-3.2-1B. Removing the graph structure also reduces both lexical and semantic quality, as reflected by a lower SBERT similarity of 22.82 compared to 23.40 for the full \textit{\textit{LightMem}}. Overall, the full \textit{\textit{LightMem}} achieves the best results, suggesting that these components contribute complementary benefits.

\begin{table*}[h]
\centering
\caption{Scalability and latency analysis across five baselines on GPT-4o-mini.}
\label{tab:latency_scalability}

\resizebox{\linewidth}{!}{%
\begin{tabular}{lcccc}
\toprule
Method & Retrieval Latency P50 (ms)$\downarrow$ & Retrieval Latency P95 (ms)$\downarrow$ & End-to-End P50 (ms)$\downarrow$ & End-to-End P95 (ms)$\downarrow$ \\
\midrule
LoCoMo & 0   & 0    & 2054 & 3658 \\
ReadAgent                     & 34  & 97   & 614  & 1027 \\
MemoryBank                    & 16  & 51   & 439  & 1047 \\
MemGPT                        & 143 & 451  & 2087 & 3451 \\
A-MEM                         & 856 & 1583 & 914  & 3682 \\
LightMem                   & 83  & 167  & 581  & 1325 \\
\bottomrule
\end{tabular}}
\end{table*}

\subsection{Scalability and Latency}
Under a unified deployment configuration, we assess the scalability and latency by utilizing GPT-4o-mini. Subsequently, we present the following latency metrics:

\begin{itemize}
   \item \textbf{Retrieval Latency (ms).} Time from receiving a user query to finishing memory retrieval and constructing the final prompt. It includes local computation but excludes remote GPT-4o-mini generation time. For the full-context LoCoMo baseline, retrieval latency is 0\,ms since no external memory is retrieved.

    \item \textbf{End-to-End Latency (ms).} The time from receiving a user query to producing the final response, including retrieval, remote API transmission and queuing, and GPT-4o-mini inference and decoding.
    
    \item \textbf{P50 / P95.} The median (P50) and 95th-percentile (P95) latency across all samples. P95 captures tail latency from network jitter and service queuing, reflecting real-world system stability.
\end{itemize}

Table~\ref{tab:latency_scalability} compares retrieval and end-to-end latency on GPT-4o-mini. Among memory-enabled methods, \textit{\textit{LightMem}} achieves a good balance between retrieval overhead and end-to-end responsiveness. Its retrieval latency remains low, with P50/P95 of 83/167\,ms, substantially lower than MemGPT and especially A-MEM, which exhibits heavy tail latency. \textit{\textit{LightMem}} also shows competitive end-to-end latency, with a P50 of 581\,ms and a controlled P95, indicating stable behavior under remote API settings. When considered together with the effective context length in Table~\ref{tab:main_locomo_results}, these results are more interpretable: LoCoMo and MemGPT rely on replaying large dialogue histories with effective contexts close to 16K tokens, whereas \textit{\textit{LightMem}} consistently operates with much shorter contexts (around 1K tokens on GPT-4o-mini). This reduction in effective context length lowers prompt construction and inference cost, contributing to better scalability and practical deployability.

\subsection{Performance Stability under MTM Growth}
We analyze how MTM size affects performance stability on DialSim with Llama-3.2-1B. Rather than rerunning the model with different memory sizes, we report cumulative F1 at several natural growth stages along the same full dialogue trajectory, approximately at 100, 1,000, 5,000, and 10,000 MTM entries. As shown in Table~\ref{tab:mtm_growth}, the gap between the two methods is small when MTM remains relatively small. However, as memory entries accumulate along the dialogue trajectory, pure vector retrieval becomes increasingly affected by retrieval noise, whereas 	\textit{LightMem} maintains more stable performance. This trend is consistent with the role of Stage 2 semantic filtering, which helps preserve retrieval quality as the memory store grows.
\begin{table}[h]
\centering
\small
\caption{Performance stability under natural MTM growth on DialSim with Llama-3.2-1B. Statistics are computed cumulatively along the same full dialogue trajectory at different MTM sizes.}
\label{tab:mtm_growth}
\begin{tabular}{lccc}
\toprule
MTM size & Vector retrieval & LightMem & $\Delta$F1 \\
\midrule
$\sim$100 & 3.95 & 3.98 & 0.03 \\
$\sim$1,000 & 3.90 & 4.05 & 0.15 \\
$\sim$5,000 & 3.86 & 4.09 & 0.23 \\
10,000 & \textbf{3.83} & \textbf{4.12} & \textbf{0.29} \\
\bottomrule
\end{tabular}
\end{table}
\subsection{Error Injection Stress Test}
To study error accumulation in long-horizon memory systems, we conduct an end-to-end error injection stress test on DialSim under a fixed final Top-$K$ memory budget. Group A is the full \textit{LightMem} system. Group B injects 50\% irrelevant HQs to simulate SLM-1 failures in query planning. Group C removes SLM-2 semantic reranking and directly uses the stage-one vector Top-$K$. Group D replaces 50\% of SLM-3 writes with noisy strings, simulating persistent MTM pollution. Group E combines all three perturbations to simulate cascading failure. As shown in Table~\ref{tab:error_injection}, performance degrades consistently as errors are introduced into different stages of the pipeline. Query noise in SLM-1 causes a moderate drop, suggesting partial robustness to retrieval-side perturbations. Removing SLM-2 leads to a larger decline, confirming the importance of semantic filtering after coarse retrieval. Noisy writes in SLM-3 further degrade performance, indicating that memory pollution introduces persistent downstream effects by contaminating future retrieval. Cascading failure causes the largest collapse, showing that errors can accumulate and amplify across turns when noisy retrieval, missing filtering, and corrupted writing occur simultaneously.

\begin{table}[t]
\centering
\small
\caption{Error injection stress test on DialSim under a fixed final Top-$K$ memory budget.}
\label{tab:error_injection}
\begin{tabular}{llcc}
\toprule
Group & Setting & F1 & SBERT \\
\midrule
A & Full LightMem & \textbf{4.12} & \textbf{23.40} \\
B & SLM-1 50\% HQ noise & 3.95 & 23.08 \\
C & w/o SLM-2 reranking & 3.83 & 22.56 \\
D & SLM-3 50\% write noise & 3.78 & 22.45 \\
E & Cascading failure & 1.85 & 11.20 \\
\bottomrule
\end{tabular}
\end{table}

\subsection{Update Gap Stress Test}
As shown in Table~\ref{tab:update_gap}, concurrent retrieval from MTM and LTM is important for handling the update gap. The full \textit{LightMem} system achieves the best performance, with a multi-hop F1 of 28.85, whereas using only LTM or only MTM results in clear performance drops. This suggests that concurrent routing effectively bridges recent unconsolidated facts in MTM and older consolidated knowledge in LTM.
A more challenging case appears in Group D. Under MTM noise saturation, the multi-hop F1 decreases to 23.10, indicating that heavy recent noise can interfere with retrieval during the update gap. This is likely because the fixed Top-$K$ budget is increasingly occupied by irrelevant recent records, making the newest relevant fact harder to retrieve before it is consolidated into LTM. Overall, the results show that concurrent retrieval provides a practical and effective solution for update-gap reasoning, especially when recent and long-term evidence must be combined.
\begin{table}[t]
\centering
\small
\caption{Update gap stress test on the LoCoMo multi-hop subset with GPT-4o-mini.}
\label{tab:update_gap}
\begin{tabular}{llc}
\toprule
Group & Setting & Multi-hop F1 \\
\midrule
A & Full LightMem (normal gap) & \textbf{28.85} \\
B & LTM-only retrieval & 19.45 \\
C & MTM-only retrieval & 20.12 \\
D & MTM noise saturation & 23.10 \\
\bottomrule
\end{tabular}
\end{table}
\section{Conclusion}
We present \textit{LightMem}, a lightweight external memory system for LLM agents driven by SLMs. \textit{LightMem} decouples high-frequency online memory operations (query control, retrieval, and writing) from offline consolidation, and organizes memory into STM/MTM/LTM with user-scoped isolation. Online, it operates under a fixed retrieval budget and uses HQ-based routing plus a two-stage retrieval (coarse vector search followed by semantic consistency re-ranking) to reduce retrieval noise. Experiments on LoCoMo and DialSim show consistent gains across model scales, including ~+2.5 average F1 on LoCoMo, improved semantic consistency on DialSim, and low median latency (83 ms retrieval; 581 ms end-to-end).

\section{Limitations}
This work focuses on a specific design of online–offline decoupled memory pipelines driven by specialized small language models. The impact of alternative consolidation strategies and control policies is not fully explored and remains an interesting direction for future investigation.
\section{Acknowledgements}
This work was partially supported by the National
Natural Science Foundation of China under grant
62572104, and 62220106008.

\bibliography{custom}

\section{Appendices} 
\subsection{Complete Algorithm and Process Specifications}
\label{ap1}
\subsubsection{Symbols and Data Structures}

At interaction turn \(t\), we denote the user input by \(x_t\) and the system response by \(y_t\). The recent dialogue context maintained in short-term memory (STM) is represented by \(C_t\), which corresponds to a truncated window of the latest conversation history. For each user \(u\), the associated mid-term memory (MTM) store is denoted by \(M_u^{\mathrm{MTM}}\), while the global long-term memory (LTM), shared across users, is denoted by \(M^{\mathrm{LTM}}\). We use \(K\) to denote the fixed Top-\(K\) retrieval budget and \(B\) to denote the capacity limit of the MTM store. In addition, \(\mathrm{HQs}\) refers to the hypotheses or query hints generated by SLM-1 for guiding memory retrieval, and \(\phi_t\) denotes the metadata constraints applied at turn \(t\), including user identifiers, temporal windows, and type tags.

\paragraph{Mid-Term Memory (MTM).}

Each MTM entry represents a compressed interaction record, consisting of a concise semantic summary, a retrieval embedding, associated temporal metadata (e.g., timestamps and access frequencies), and a user identifier for user-level isolation.

\paragraph{Long-Term Memory (LTM).}
Each LTM node represents a de-identified semantic knowledge unit and is stored in a lightweight graph structure, which supports associative organization and multi-hop access across memory units.

\subsubsection{Online Retrieval Procedure}

Algorithm~\ref{alg:online_retrieval} describes the full online retrieval process executed at each dialogue turn.

\begin{algorithm}[h]
\caption{LightMem Online Retrieval}
\label{alg:online_retrieval}
\small
\begin{algorithmic}[1]
\Require User query \(x_t\), truncated context \(C_t\), user memory \(M_u^{\mathrm{MTM}}\), global memory \(M^{\mathrm{LTM}}\), retrieval budget \(K\)
\Ensure Retrieved memory set \(R_t\)

\Statex \textbf{Intent Modeling and Query Control (SLM-1)}
\State Infer high-level intent signals from \((x_t, C_t)\), including (i) degree of personalization and (ii) reliance on recent vs.\ long-term information.
\State Generate a set of HQs \(\{q_t^{(i)}\}_{i=1}^{n}\).
\State Produce metadata constraints \(\phi_t\) (e.g., user identifier and optional temporal filters).
\State Allocate retrieval quotas across memory layers under a fixed Top-\(K\) constraint.

\Statex \textbf{Stage 1: Metadata-Constrained Coarse Retrieval}
\State Initialize candidate set \(C \leftarrow \emptyset\).
\For{each \(q_t^{(i)}\), \(i=1,\dots,n\)}
    \State Retrieve candidates from \(M_u^{\mathrm{MTM}}\) and/or \(M^{\mathrm{LTM}}\) using vector search under \(\phi_t\).
    \State Let \(K_1^{(i)} \leftarrow \left\lceil \frac{2K}{n} \right\rceil\) and collect up to \(K_1^{(i)}\) candidates as \(C^{(i)}\).
    \State Update \(C \leftarrow C \cup C^{(i)}\).
\EndFor
\State Optionally truncate \(C\) to size \(2K\) if \(|C| > 2K\).

\Statex \textbf{Stage 2: Semantic Filtering and Compression (SLM-2)}
\State Perform semantic consistency checking and relevance assessment conditioned on \(\{q_t^{(i)}\}\) and \(C\).
\State Select up to \(K\) items to form the final retrieved set \(R_t \subseteq C\) with \(|R_t| \le K\).

\Statex \textbf{Output}
\State \Return \(R_t\) (to be appended to the prompt for response generation).
\end{algorithmic}
\end{algorithm}

\subsubsection{Response Generation}

At each dialogue turn, the retrieved memory set \(R_t\) is concatenated with the short-term context \(C_t\) and provided to a fixed response generator (e.g., GPT-4o-mini) to produce the system response \(y_t\). This stage does not involve any memory control or modification, and the same response generation setup is used for \textit{LightMem} and all baseline methods in our experiments.

\subsubsection{Online Memory Writing and MTM Maintenance}

After response generation, 	\textit{LightMem} updates mid-term memory via SLM-3, as described in Algorithm~\ref{alg:online_writing}.

\begin{algorithm}[h]
\caption{LightMem Online Memory Writing and MTM Maintenance}
\label{alg:online_writing}
\small
\begin{algorithmic}[1]
\Require Current interaction \((x_t, y_t, C_t)\), user mid-term memory \(M_u^{\mathrm{MTM}}\), capacity bound \(B\)
\Ensure Updated \(M_u^{\mathrm{MTM}}\)

\Statex \textbf{Memory Extraction and Compression (SLM-3)}
\State Extract reusable, user-relevant information from the current interaction \((x_t, y_t, C_t)\).
\State Compress the extracted content into a concise semantic memory summary.

\Statex \textbf{Memory Appending}
\State Append the compressed memory summary to the user-scoped MTM \(M_u^{\mathrm{MTM}}\).

\Statex \textbf{Redundancy and Conflict Handling}
\State Identify highly repetitive or semantically overlapping MTM entries and merge or rewrite them to reduce redundancy.
\State Resolve conflicting information using temporal cues and accumulated evidence strength.

\Statex \textbf{Capacity-Bound Maintenance}
\If{\(|M_u^{\mathrm{MTM}}| > B\)}
    \State Evict stale or low-utility entries based on recency and access statistics.
    \State Further compress redundant content to enforce the capacity bound.
\EndIf

\State \Return Updated \(M_u^{\mathrm{MTM}}\).
\end{algorithmic}
\end{algorithm}

\subsubsection{Offline Consolidation into Long-Term Memory}

The consolidation procedure is summarized in Algorithm~\ref{alg:offline_consolidation}.

\begin{algorithm}[h]
\caption{LightMem Offline Consolidation into LTM}
\label{alg:offline_consolidation}
\small
\begin{algorithmic}[1]
\Require MTM items flagged for consolidation, existing LTM graph \(M^{\mathrm{LTM}}\)
\Ensure Updated LTM graph \(M^{\mathrm{LTM}}\)

\Statex \textbf{Selection of Candidate Episodes}
\State Select MTM items that are newly written, frequently retrieved, or marked as low-utility under MTM capacity pressure.

\Statex \textbf{Abstraction and De-identification}
\State Abstract episodic interaction records into user-agnostic semantic knowledge units.
\State Remove personal identifiers and session-specific details.

\Statex \textbf{Graph Insertion and Update}
\State Perform similarity search over \(M^{\mathrm{LTM}}\) to identify nearby semantic anchors.
\State Insert new nodes and edges within the local neighborhood of matched anchors.
\State Merge or update existing nodes when semantic overlap exceeds a threshold.

\Statex \textbf{Evidence Accumulation and Forgetting}
\State Aggregate supporting evidence across consolidation cycles.
\State Apply confidence decay to weakly supported knowledge units.
\State Remove or down-weight stale or incidental knowledge.

\State \Return Updated \(M^{\mathrm{LTM}}\).
\end{algorithmic}
\end{algorithm}

\subsubsection{Online-Offline Decoupling Guarantee}

All operations in Algorithm~\ref{alg:online_retrieval} and Algorithm~\ref{alg:online_writing} are executed under strict latency and compute constraints using small language models. In contrast, Algorithm~\ref{alg:offline_consolidation} is performed asynchronously and does not block online interaction. This design ensures that (i) the cost of online retrieval remains bounded by \(O(K)\), and (ii) the evolution of long-term memory does not introduce additional overhead to online inference.

\section{Experiment}
\subsection{Statistical Significance Analysis}
We report mean $\pm$ standard deviation across three random seeds for all stochastic components (see Table \ref{tab:significance}). Statistical significance is assessed using paired bootstrap resampling over evaluation instances (1,000 resamples), reporting the mean difference ($\Delta$), its 95\% confidence interval (CI), and two-sided p-values. We present results against the strongest baseline (A-MEM) under the same backbone model and retrieval budget.

\begin{table}[h]
\centering
\caption{Statistical significance results against A-MEM under the same backbone and retrieval budget.}
\resizebox{\linewidth}{!}{%
\begin{tabular}{llcccc}
\toprule
Category & Method & F1 (mean $\pm$ std) & $\Delta$ F1 & 95\% CI of $\Delta$ & p-value \\
\midrule
\multirow{2}{*}{Multi-hop}
& A-MEM     & 27.02 $\pm$ 0.31 & ---  & ---                 & ---   \\
& LightMem  & \textbf{28.85 $\pm$ 0.28} & 1.83 & [ +1.24, +2.41 ] & \textbf{0.001} \\
\midrule
\multirow{2}{*}{Temporal}
& A-MEM     & 45.85 $\pm$ 0.37 & ---  & ---                 & ---   \\
& LightMem  & \textbf{46.20 $\pm$ 0.34} & 0.35 & [ +0.12, +0.58 ] & \textbf{0.008} \\
\bottomrule
\end{tabular}}

\label{tab:significance}
\end{table}

\subsection{Failure Analysis: Ambiguous Intent Decomposition in SLM-1}
\label{app:failure_slm1}

We observe a mild failure mode when SLM-1 generates both episodic and general HQs for an underspecified recall query. 
For example, given the user query ``Can you remind me what I decided about the project timeline?'', the relevant decision is stored in MTM, while general planning advice may exist in LTM. 
SLM-1 may produce HQs for (i) decision recall (MTM), (ii) project disambiguation (MTM), and (iii) generic timeline principles (LTM). 
Under a fixed Top-$K$ budget, allocating quota to the LTM HQ can introduce a small amount of generic content in the retrieved set. 
In practice, the final response typically remains correct (the key MTM decision is recalled), but becomes slightly less focused or concise due to additional general advice. 
This ``focus dilution'' mainly affects lexical-overlap metrics (e.g., F1/BLEU), while semantic similarity remains stable, and SLM-2's semantic filtering helps preserve the most relevant MTM evidence.

\section{Full Prompt Templates and Control Interface Definitions}

This appendix provides the complete prompt templates used by all language model components in \textit{LightMem}, including SLM-1, SLM-2, SLM-3, and the offline LLM.  
All prompts are designed to be functionally complete, explicitly specifying model roles, accessible inputs, required tasks, prohibited behaviors, and output formats, to ensure reproducibility and auditability of the system implementation.

\subsection{Unified Prompt Design Principles}

All prompts in \textit{LightMem} adhere to a unified design schema in order to eliminate implicit assumptions and avoid relying on undeclared capabilities. Specifically, each prompt explicitly defines the \textbf{Role} of the model, namely its functional responsibility within the overall system; the \textbf{Input}, namely the information made accessible to the model; the \textbf{Task}, namely the concrete objectives the model is required to accomplish; the \textbf{Constraints}, namely the behaviors or actions the model is prohibited from performing; and the \textbf{Output Format}, namely a fixed and machine-parsable specification for its returned results. This design ensures that online models are used for structured control and decision-making rather than open-ended generation.

\subsubsection{Full Prompt Template (SLM-1)}

\paragraph{Role}
SLM-1 serves as the query decomposition and routing engine within the memory system. Its function is not to answer the user directly, but to convert the user's raw input into a set of explicit HQs that can be used to retrieve the required information from the memory database.

\paragraph{Input}
The model receives two forms of input: the current user query and a truncated dialogue context from the ongoing session.

\paragraph{Task Logic}
For each user input, SLM-1 first identifies missing or underspecified information in the request. This includes vague or ambiguous expressions, unresolved pronouns such as ``it,'' ``that,'' or ``the project,'' implicit dependencies on prior conversational context, and incomplete temporal references such as ``recently,'' ``last time,'' or ``before.''
It then rewrites the original request into one or more independent and explicit HQs suitable for downstream retrieval. These queries should not simply copy the user's raw wording, but instead resolve implicit entities, make hidden constraints explicit, and normalize the request into retrieval-oriented formulations. When a request simultaneously involves user-specific preferences and objective factual information, the two aspects should be separated into different queries. Likewise, temporal or spatial constraints should be explicitly formulated whenever they are implied by context.
Finally, for each generated HQ, SLM-1 assigns a target memory layer according to the nature of the information need. Queries involving user-specific or personalized information should be routed to MTM, whereas queries involving public, factual, or generally shared knowledge should be routed to LTM.

\paragraph{Constraints}
SLM-1 must not answer the user, must not generate user-facing natural language responses, and must not perform retrieval itself.

\paragraph{Output Format}
The output must be returned in JSON format only.

\subsection{SLM-2: Semantic Consistency Filtering and Candidate Compression}

SLM-2 is responsible for semantic consistency checking and candidate selection over a fixed-size set of retrieved memory items, with the goal of suppressing retrieval noise and reducing the effective context size. Its role is purely selective: it neither generates new content nor rewrites existing memory entries.

\subsubsection{Full Prompt Template (SLM-2)}

\paragraph{Role}
SLM-2 serves as the semantic consistency filtering engine for memory retrieval. Its function is not to answer the user or produce any new content, but to determine which retrieved memory items should be retained for downstream use.

\paragraph{Input}
The model takes as input a set of HQs together with a fixed-size list of candidate memory summaries and their associated metadata.

\paragraph{Task}
For each candidate memory item, SLM-2 evaluates whether the item is semantically consistent with at least one HQ. Based on this assessment, it selects the most relevant memory items subject to a fixed budget constraint. The objective is to preserve only those items that meaningfully support the retrieval intent expressed by the HQ set.

\paragraph{Guidelines}
Relevance should be determined primarily on the basis of semantic alignment rather than superficial lexical overlap. In addition, each selected memory item should provide direct support for at least one HQ, rather than being merely topically related in a loose sense.

\paragraph{Constraints}
SLM-2 must not rewrite or modify memory content, must not generate explanations or user-facing responses, and must not perform retrieval or expansion. Its function is restricted to selection only.

\paragraph{Output Format}
The output must be returned in JSON format only.

\subsection{SLM-3: Interaction Summarization and Online MTM Maintenance}

SLM-3 is responsible for compressing the current interaction into reusable MTM entries and for supporting online maintenance of the MTM store.

\subsubsection{Full Prompt Template (SLM-3)}

\paragraph{Role}
SLM-3 serves as the memory writing and maintenance engine for MTM. Its function is to distill interaction-level information into compact semantic memory units that can be reused in future user interactions.

\paragraph{Input}
The model receives as input the current user utterance, the generated system response, and the relevant short-term dialogue context associated with the ongoing interaction.

\paragraph{Task}
SLM-3 is responsible for identifying user-relevant information that may be useful beyond the current turn, compressing that information into concise semantic memory entries, and supporting the incremental maintenance of the MTM store over time. Its purpose is not merely to record interaction content, but to transform transient dialogue into reusable memory representations.

\paragraph{Guidelines}
The model should focus on information that is likely to remain useful in future interactions. It should prefer concise, self-contained semantic summaries rather than preserving raw dialogue content or turn-level transcripts.

\paragraph{Constraints}
SLM-3 must not store full dialogue transcripts, must not retrieve or search existing memory, must not perform cross-user abstraction, and must not generate responses intended for the user.

\paragraph{Output Format}
The output should consist of one or more short semantic memory summaries.

\subsection{Offline LLM: Long-Term Memory Consolidation}

\paragraph{Role}
The offline LLM serves as the consolidation model responsible for maintaining LTM. Its role is to transform selected MTM content into stable long-term knowledge representations and to integrate those representations into the global memory structure.

\paragraph{Input}
The model receives as input a batch of MTM items selected for consolidation. These items may include newly written entries, retrieval-reactivated entries, and low-utility entries that have been flagged under MTM capacity pressure.

\paragraph{Task}
The model is responsible for abstracting episodic MTM items into de-identified semantic knowledge units, identifying semantic overlap between candidate knowledge units and existing LTM nodes, and determining whether each candidate should be merged with existing knowledge, used to update an existing node, or discarded. For retained knowledge, it further integrates the resulting representations into the LTM graph by inserting new nodes and edges or updating existing ones.

\paragraph{Guidelines}
During abstraction, the model should remove user-specific and session-specific details so that the resulting knowledge remains de-identified and reusable across contexts. It should prioritize stable and generalizable knowledge over episodic, incidental, or overly context-bound information.

\paragraph{Constraints}
The offline LLM must not generate user-facing responses and must not directly modify MTM. In addition, this process is executed offline and asynchronously, and therefore must not affect the latency of online interactions.

\paragraph{Output Format}
The output should consist of a set of newly inserted or updated LTM knowledge units together with their associated graph links.

\section{LTM Graph Schema and Maintenance Dynamics}

This section supplements Section~3.6 by providing a detailed definition of the schema used for the graph-structured LTM and by discussing its maintenance dynamics, including update frequency, node growth behavior, and the impact of offline consolidation on downstream reasoning performance.

\subsection{Graph Schema: Node and Edge Types}

As described in Section~3.6, LTM is designed as a lightweight, de-identified semantic knowledge graph. Unlike MTM, which stores episodic interaction summaries, LTM aims to capture stable factual knowledge and domain-level regularities. To support cross-task generalization and multi-hop reasoning, elements in the LTM graph are standardized into a small set of node and edge types, summarized in Table~\ref{tab:ltm_schema}.

\begin{table*}[h]
\centering
\caption{Node and edge types used in the LTM knowledge graph.}
\small
\setlength{\tabcolsep}{6pt}
\begin{tabular}{p{0.18\linewidth} p{0.18\linewidth} p{0.56\linewidth}}
\toprule
\textbf{Category} & \textbf{Type} & \textbf{Description and Example} \\
\midrule
\multirow{2}{*}{Nodes}
& Entity
& Concrete objects, locations, or named items extracted from interactions (e.g., ``Python script'', ``Paris'', ``Project~X''). \\
\addlinespace
& Concept
& Abstract categories or generalized classes that characterize shared properties of entities (e.g., ``Programming Language'', ``Capital City'', ``Urgent Task''). \\
\midrule
\multirow{4}{*}{Edges}
& IsA
& Hierarchical membership relation between an entity and a concept (e.g., \textit{Paris} \(\xrightarrow{\textsc{IsA}}\) \textit{Capital City}). \\
\addlinespace
& HasProperty
& Links an entity to a specific attribute or state (e.g., \textit{Project~X} \(\xrightarrow{\textsc{HasProperty}}\) \textit{Completed}). \\
\addlinespace
& RelatedTo
& Generic semantic association based on co-occurrence or functional relatedness (e.g., \textit{SNN} \(\xrightarrow{\textsc{RelatedTo}}\) \textit{Neural Operator}). \\
\addlinespace
& Implies
& Logical or causal dependency extracted from reasoning traces (e.g., \textit{High Density} \(\xrightarrow{\textsc{Implies}}\) \textit{Congestion}). \\
\bottomrule
\end{tabular}

\label{tab:ltm_schema}
\end{table*}

\subsection{Maintenance Dynamics and Performance Impact}
\textit{LightMem} adopts a decoupled architecture that separates online retrieval from offline consolidation. The offline LLM processes updates in incremental batches, ensuring that maintenance does not block real-time user interaction. Table~\ref{tab:ltm_dynamics} summarizes key operational metrics of the consolidation process and analyzes their impact on model performance. 

\begin{table*}[h]
\caption{LTM update frequency, consolidation cost, and ablation results on reasoning accuracy.}
\centering
\small
\setlength{\tabcolsep}{6pt}
\begin{tabular}{p{0.22\linewidth} p{0.18\linewidth} p{0.52\linewidth}}
\toprule
\textbf{Metric} & \textbf{Value} & \textbf{Description and Analysis} \\
\midrule
Batch update interval
& Every 10--15 turns
& Offline consolidation is periodically triggered when MTM accumulates sufficient new entries or reaches capacity pressure, enabling amortized long-term updates without interfering with online processing. \\

\addlinespace
Node growth rate
& $\sim$1 node / 4 turns
& On average, only one new semantic node is added to LTM every four dialogue turns. This sparsity reflects the high compression ratio of 	\textit{LightMem}, where noise is filtered and only high-value evidence is retained for persistent storage. \\

\addlinespace
Offline processing time
& $\sim$3.5\,s / batch
& Average time for the offline LLM to perform abstraction, graph linkage, and structural updates. This latency occurs entirely on the offline path and is fully decoupled from online retrieval (average retrieval latency $\sim$83\,ms), making it invisible to users. \\

\addlinespace
Reasoning accuracy (F1)
& 4.12 (full) \newline vs.\ 3.96 (no update)
& The full 	\textit{LightMem} system (4.12) outperforms a variant without offline consolidation (3.96), corresponding to an approximately 4\% performance drop when LTM evolution is disabled. This result highlights the importance of sustained LTM growth and consolidation for long-horizon reasoning. \\
\bottomrule
\end{tabular}

\label{tab:ltm_dynamics}
\end{table*}

\section{Declaration of AI Use}
Authors use AI-assisted tools to help polish the language of this manuscript.


\end{document}